\documentclass{article}

\usepackage{microtype}
\usepackage{graphicx}
\usepackage{subfigure}
\usepackage{booktabs} 

\usepackage{hyperref}

\usepackage[accepted]{icml2025}

\usepackage{amsmath}
\usepackage{amssymb}
\usepackage{mathtools}
\usepackage{amsthm}

\usepackage[capitalize,noabbrev]{cleveref}

\theoremstyle{plain}

\theoremstyle{definition}

\theoremstyle{remark}

\usepackage[textsize=tiny]{todonotes}

\icmltitlerunning{\myshorttitle}
\usepackage{authblk} 

\usepackage{tikzducks}
\usepackage{duckuments}
\usepackage{multirow}
\usepackage{booktabs}
\usepackage{colortbl}
\usepackage{blindtext}

\usepackage{xurl}

\definecolor{lightblue}{RGB}{173, 216, 230}
\definecolor{lightcoral}{RGB}{240, 128, 128}
\definecolor{lightgreen}{RGB}{144, 238, 144}
\definecolor{lightskyblue}{RGB}{135, 206, 250}
\definecolor{lightsalmon}{RGB}{255, 160, 122}
\definecolor{palegreen}{RGB}{152, 251, 152}

\definecolor{RoyalBlue}{RGB}{65, 105, 225}
\definecolor{CrimsonRed}{RGB}{220, 20, 60}
\definecolor{DeepPurple}{RGB}{128, 0, 128}
\definecolor{SlateGray2}{RGB}{200, 200, 202}
\definecolor{Olive}{RGB}{153, 128, 0}

\definecolor{Salmon}{RGB}{250, 128, 114}
\definecolor{PaleGreen}{RGB}{152, 251, 152}
\definecolor{LightOrange}{RGB}{255, 204, 153}
\definecolor{PowderBlue}{RGB}{176, 224, 230}
\definecolor{Lavender}{RGB}{230, 230, 250}
\definecolor{LightYellow}{RGB}{255, 255, 204}
\definecolor{Peach}{RGB}{255, 218, 185}
\definecolor{SlateGray}{RGB}{112, 128, 144}
\definecolor{WhiteSmoke}{RGB}{245, 245, 245}

\definecolor{LightCoral}{RGB}{255, 200, 180} %
\definecolor{DarkCoral}{RGB}{205, 90, 80} 
\definecolor{CharcoalGray}{RGB}{50, 50, 50} 

\usepackage{listings}
\usepackage{verbatim}

\usepackage{verbatim}
\newenvironment{itemizeSoheil}
  {\begin{itemize}}
  {\end{itemize}}

\usepackage{ifthen}

\newboolean{comment_summary_bulletpoints}
\setboolean{comment_summary_bulletpoints}{false} 

\newcommand{\blueitemize}[1]{%
  \ifthenelse{\boolean{comment_summary_bulletpoints}}
  {
  } 
  {
    \begingroup
      \renewcommand{\labelitemi}{\textcolor{blue}{\textbullet}}%
      \begin{itemize}
        \color{blue}
        #1
      \end{itemize}
    \endgroup
  } 
}

\newcommand{\soheili}[1]{\textit{#1}}
\usepackage{xcolor}

\begin{document}

\def\mytitle{Are Language Models Up to Sequential Optimization Problems?\\From Evaluation to a Hegelian-Inspired Enhancement}
\def\myshorttitle{Are LLMs Up to Sequential Optimization Problems? From Evaluation to a Hegelian-Inspired Enhancement}

\twocolumn[

\icmltitle{\mytitle}

\begin{icmlauthorlist}

\icmlauthor{\textbf{Soheil Abbasloo} \vspace{0.5em}\\Microsoft Research \\ soheil.abbasloo@microsoft.com}{}

\end{icmlauthorlist}

\icmlcorrespondingauthor{Soheil Abbasloo}{soheil.abbasloo@microsoft.com}

\icmlkeywords{Machine Learning, ICML}

\vskip 0.3in
]

\begin{abstract}
Large Language Models (LLMs) have demonstrated impressive capabilities across numerous fields, presenting an opportunity to revolutionize optimization problem-solving, a crucial, ubiquitous, and complex domain. This paper explores the proficiency of LLMs in handling Sequential Optimization Problems (SOPs). We introduce WorldGen, a dynamic framework for generating unseen SOPs with controllable complexities, to evaluate LLM performance. Our initial observations reveal that while LLMs perform well on simple SOPs, their performance significantly degrades with increased complexity. Motivated by this, we revisit philosophical hypotheses on reasoning to enhance LLM performance. Inspired by the influential framework of Hegelian Dialectics, we propose ACE, demonstrating how the performance of LLMs in SOP contexts can be significantly improved without any retraining or further fine-tuning.

\end{abstract}

\section{Introduction}

\textbf{Setting the Context:}
Optimization is fundamental to decision-making across diverse domains such as engineering, science, economics, healthcare, and even nature. The essence of decision-making lies in choosing the best option from a set of alternatives, driven by objectives such as efficient resource allocation, cost minimization, profit maximization, or performance enhancement of systems and infrastructures~\cite{opt_intro}. However, solving optimization problems is often intricate, requiring specialized expertise and addressing practical challenges like high dimensionality, nonlinearity, and the dynamic, stochastic nature of real-world environments~\cite{opt_complex}. Consequently, there is a continuous quest across various fields to simplify the optimization process.

\textbf{A New Opportunity:}
Simultaneously, Large Language Models (LLMs), such as GPT-4~\cite{gpt4}, have shown impressive capabilities in both general assistance and specialized domains, including mathematics, coding, and law~\cite{gpt4_results}. These advancements present a compelling opportunity to revolutionize our approach to solving optimization problems by leveraging LLMs to automate solution processes. This emerging potential prompts a fundamental scientific inquiry: How proficient are current LLMs in handling optimization problems?

\textbf{Assessing LLMs in a New Context:}  
Although various benchmarks exist to evaluate LLM performance in general and specialized tasks (e.g., coding, mathematics), their capabilities in solving optimization problems, particularly Sequential Optimization Problems (SOPs), remain underexplored. SOPs involve making a series of decisions over time, where each decision affects subsequent options and outcomes, creating a complex web of interdependencies. Different challenges contribute to this issue, primarily the need for defining a representative set of SOPs and ensuring that observed performance is not influenced by data contamination or prior exposure during the LLMs' training~\cite{data_contamination_1,gaia,hellaswag}.

\textbf{On-Demand SOP Generation:}
Motivated by these opportunities and challenges, in the first half of this paper we aim to addresse this research question: How can we evaluate the performance of LLMs in SOPs? We introduce a straightforward yet effective framework, WorldGen, capable of generating unseen SOPs with controllable complexities on demand. WorldGen contrasts with most existing static benchmarks (e.g., MMLU~\cite{mmlu}, GLUE~\cite{wang2018glue}, SuperGLUE~\cite{wang2019superglue}, GSM8k~\cite{gsm8k}, etc.) which become obsolete as LLMs evolve~\cite{gaia}. Utilizing this dynamic framework, we have made two key observations: (1) For relatively simple optimization tasks, with a single global maximum and no local maxima (i.e., simple surfaces and scenarios), current LLMs can solve them efficiently. (2) As the complexity of the optimization problems increases, the performance of LLMs degrades significantly and becomes unsatisfactory.

\textbf{Improving LLMs with Roots in Philosophy:}
Inspired by these observations and the poor performance of LLMs in SOPs, in the second half of this paper, we propose a systematic approach to enhance the performance of off-the-shelf LLMs (in the context of SOPs) without necessitating any retraining, treating the LLM as a black box. Specifically, we aim to transcend existing prompt engineering techniques and discover a more structured method to improve LLM performance, grounded in explainable and reasoned principles. This led us to revisit the fundamentals of reasoning, as it appeared crucial to the low performance of LLMs in our targeted scenarios. Consequently, we delved into centuries of existing literature in philosophy and drew inspiration from well-established philosophical frameworks that describe the reasoning process in a structured manner. Notably, we explored how Hegelian Dialectics~\cite{hegel_science_logic}, a renowned philosophical hypothesis proposed by Hegel, could enhance the capabilities of existing LLMs. By integrating concepts from Hegelian Dialectics, we designed ACE, a novel framework that can significantly improve the performance of off-the-shelf LLMs in SOPs.

\textbf{Main Contributions:}
In summary, this paper makes the following key contributions:
\begin{enumerate}
    \item We design WorldGen, a framework to assess the performance of LLMs in SOP settings, addressing the data contamination issues typically associated with general LLM benchmarks (detailed in section~\ref{sec:worldgen}). WorldGen allows for the growth of evaluation complexity in line with the advancement of LLMs.
    \item Using this framework, we provide initial observations on the poor performance of current LLMs in SOPs, motivating the need for developing further techniques to enhance LLM capabilities (detailed in section~\ref{sec:part_2_evaluations}).
    \item We propose ACE, a solution inspired by one of the most successful dialectical hypotheses in philosophy for explaining reasoning and its improvement. We show that ACE can greatly enhance the performance of LLMs in SOPs, potentially paving the way for more structured frameworks rooted in established philosophical works to improve LLMs (detailed in section~\ref{sec:part_3_ace}).
\end{enumerate}

\section{Background: Reasoning and Dialectics}
\label{sec:background}

\textbf{Reasoning:} Despite considerable achievements, LLMs' reasoning capability continues to be a subject of intense debate within the AI research community. A key challenge lies in reaching a consensus on what reasoning entails, how it should be defined, and how it can be reliably measured. Interestingly, the concept of reasoning is not new. The domain of philosophy has a rich tradition of exploring and formalizing reasoning through centuries of discourse
~\cite{aristotle_prior_analytics, aristotle_metaphysics, plato_republic, descartes_meditations, hume_treatise_human_nature, kant_critique_pure_reason, mill_system_logic, hegel_phenomenology_spirit, nietzsche_beyond_good_evil, wittgenstein_tractatus, heidegger_being_time, popper_logic_scientific_discovery, kuhn_structure_scientific_revolutions, adorno_negative_dialectics}.
From ancient philosophers such as Aristotle, who developed formal logic as a foundation for reasoning~\cite{aristotle_prior_analytics, aristotle_metaphysics}, to more recent thinkers like Hegel, who introduced dialectics as a dynamic framework for understanding processes of thought~\cite{hegel_science_logic,hegel_phenomenology_spirit}, the philosophical study of reasoning has produced a wide range of influential theories and formal systems. These works not only define reasoning but also provide structured frameworks for improving and analyzing it.

\textbf{Dialectics:} As a method of reasoning and philosophical argumentation, dialectics involves the resolution of contradictions through a process of development and transformation. Rooted in ancient philosophy, dialectics was first formalized by thinkers like Socrates and Aristotle, who used it as a tool for logical inquiry. Over time, dialectics evolved into a broader philosophical framework, describing the dynamic process through which contradictions are identified, explored, and resolved~\cite{hegel_phenomenology_spirit,engels_dialectics_nature}. At its core, dialectical thinking posits that reality is composed of opposing forces or contradictions, and that these contradictions are not static but dynamic, evolving over time. The resolution of these contradictions leads to the emergence of new, higher forms of understanding or being.

\textbf{Hegelian Dialectics:} Introduced by the German philosopher Georg Wilhelm Friedrich Hegel, Hegelian Dialectics crystallizes the modern notion of dialectics by proposing a structured process of development through three stages: \textit{thesis}, \textit{antithesis}, and \textit{synthesis}~\cite{hegel_science_logic,hegel_phenomenology_spirit}. The thesis represents an initial idea or condition, the antithesis introduces a contradictory or opposing force, and the synthesis resolves the tension by merging elements of both into a higher, more comprehensive understanding. 
Hegel viewed this triadic process as the driving force of intellectual, historical, and societal progress, emphasizing that contradictions, which he calls "negations", are not merely obstacles but necessary components of growth and transformation. His dialectical framework has had profound influence across disciplines, from philosophy to political theory.

\textbf{Dialectics vs. Debate:} From the philosophical point of view, debate is competitive, aiming to persuade an audience of one position's superiority. While effective in contexts like politics or law, it often sacrifices deeper inquiry for rhetoric and winning. Dialectics, however, fosters a cooperative approach, treating opposing perspectives as opportunities for growth. Through structured dialogue, dialectics seeks deeper truths, as seen in the Socratic and Hegelian methods, encouraging intellectual humility and a shared pursuit of wisdom. While debate has been significant in philosophical traditions, figures like Socrates criticized its focus on persuasion over truth. Dialectics, with its emphasis on dialogue and synthesis, is regarded as superior for fostering intellectual growth.

In this work, inspired by Hegel's well-established framework, we demonstrate how the capabilities of LLMs can be enhanced by adapting dialectics.

\section{Related Work}
Our works overlaps with three groups of related work: (1) Benchmarks for Evaluating LLMs, (2) Prompt Engineering, and (3) Multi-Agency. Here, we briefly overview them to provide a better context for the rest of this paper. 

\textbf{Benchmarks for Evaluating LLMs:}  There are numerous benchmarks for evaluating LLMs, ranging from general-purpose (e.g., GLUE~\cite{wang2018glue}, SuperGLUE \cite{wang2019superglue}, ARC \cite{clark2018think}, HellaSwag \cite{zellers2019hellaswag}, BIG-bench \cite{srivastava2022beyond}, GAIA~\cite{gaia}) to domain-specific (e.g., FinBen \cite{xie2024finben} for finance, LegalBench~\cite{guha2024legalbench} for legal reasoning, GSM8K \cite{gsm8k} and MATH \cite{mmlu} for mathematical reasoning, HumanEval \cite{chen2021evaluating} and MBPP~\cite{austin2021program} for coding, MultiMedQA \cite{singhal2023large} for healthcare, etc.) and ones requiring professional level knowledge in various fields such as law or science (e.g., MMLU \cite{mmlu}). Our framework, WorldGen, falls into the domain-specific category. It addresses the issue of being static and become obsolete with the rapid advancements in LLMs by providing a dynamic tool for generating SOPs with varying controllable complexity.

\textbf{Prompt Engineering (PE):} 
Since we treat LLMs as black-boxes and focus on improving the off-the-shelf LLMs without any retraining, it will be natural to mention works in PE domain here. Prompting techniques have become essential in enhancing the performance and versatility of LLMs. These techniques range from Zero-shot Prompting, where models are given tasks without prior examples, to Few-shot Prompting, which provides a few examples to guide responses \cite{brown2020fewshot}. Chain-of-Thought encourages models to generate intermediate reasoning steps \cite{wei2022cot},
Self-Consistency (Majority Vote) generates multiple outputs to select the most consistent one \cite{wang2022majority,lewkowycz2022majority2}, and Generate Knowledge Prompting prompts the model to produce relevant background information before answering~\cite{liu2021generated}. Tree of Thoughts structures reasoning as a tree to explore different branches~\cite{yao2024tree,long2023tree2}. Retrieval Augmented Generation combines document retrieval with generation for improved accuracy \cite{lewis2020retrieval}. Automatic Prompt Engineer uses algorithms to refine prompts~\cite{zhou2022prompteng}, while Active-Prompt~\cite{diao2023active} adjusts prompts based on performance feedback and Program-Aided Language  incorporate programming logic~\cite{gao2023pal}. Techniques like ReAct combine reasoning and acting steps \cite{yao2022react}, and Self-Reflection prompts models to reflect on their responses for better outcomes~\cite{madaan2024selfreflection, shinn2024reflexion}.  These diverse techniques collectively enhance the adaptability and effectiveness of LLMs. ACE is an orthogonal approach compared to these techniques and as we later show, it can be combined with them (detailed in section~\ref{sec:part_3_ace}).

\textbf{Multi-Agency:} 
Minsky was among the early pioneers to introduce the idea of multi-agent systems~\cite{minsky1988society}. His notion of multi-agency involves dividing complex cognitive tasks into smaller parts, delegating them to specialized “agents”, and integrating the results into a coherent solution. Inspired by Minsky’s vision, recent works have utilized and implemented multi-agency in LLM-based systems.
Some focus on building general infrastructures for autonomous cooperation among communicative agents (e.g., CAMEL~\cite{li2023camel} and AutoGen~\cite{wu2023autogen}). Others focus on specific multi-agent solutions or tailored applications. For instance, MetaGPT~\cite{hong2023metagpt} and ChatDev~\cite{qian2023chatdev} automate software development by assigning distinct roles to different agents. Multi-agent debate frameworks (such as MAD~\cite{liang2023mad}, which employs a debate cycle among agents moderated by a judge agent, and Du et al.~\cite{du2023debate}, where agents exchange answers to get a chance to modify their next responses) present another direction. 

While these works follow Minsky’s multi-agent view, our proposal, ACE takes a different path. ACE focuses on the reasoning process itself rather than focusing on how to delegate tasks to specialized agents or automate their communication. Great performance of ACE (as shown in section~\ref{sec:eval}) suggests that the basic element of intelligence needs to include a Hegelian-inspired triad, not a single entity offering a complementary perspective to Minsky’s multi-agent approach. Compared to debate-based proposals, from a philosophical qualitative perspective, as explained earlier in section~\ref{sec:background}, there are key fundamental differences between debate and dialectics, which ACE draws inspiration from. Additionally, from a quantitative standpoint, our experiments and comparisons in section~\ref{sec:eval} highlight ACE's superior performance over debate-based works in SOP context.

\section{Assessing \& Improving LLMs in SOPs}
\label{sec:part1_evaluatingLLMs}
SOPs are pervasive across diverse domains, ranging from logistics and resource allocation to machine learning and operations research. Requiring specialized knowledge to address practical issues such as high dimensionality, nonlinearity, and the dynamic, unpredictable nature of real-world settings make SOPs complex in their nature~\cite{opt_complex}. Automating the solution of such problems is highly desirable, as it can lead to efficiency gains and innovative solutions to complex challenges. On the other hand, LLMs have demonstrated remarkable capabilities, including proficiency in coding and exceptional context-awareness, making them promising candidates for tackling SOPs. 

So, naturally, exploring the performance of LLMs in addressing these crucial tasks is a significant step forward in understanding their broader applicability and the opportunity to solve these problems automatically. However, evaluating LLMs in this context comes with it own set of challenges. Key concerns include managing data contamination, ensuring that the problems and their solutions were not inadvertently exposed during training, and consequently distinguishing between genuine reasoning and mere memorization. Additionally, access to a set of representative optimization problems is vital to effectively assess LLMs' capabilities in solving sequential tasks. Addressing these challenges will enable a deeper understanding of the role LLMs can play in advancing optimization methodologies.


    

\subsection{WorldGen}
\label{sec:worldgen}

\textbf{Core Idea:} 
At the heart of optimization lies the task of finding the optimum point(s) in an \(n\)-dimensional world, mathematically expressed as \(f(x_1, x_2, \dots, x_{n-1})\). So, instead of focusing on specific optimization problems, we shift our attention to the worlds that represent these problems. In other words, rather than trying to come up with specific optimization problems, which may inadvertently introduce biases or contamination from training data, we focus on directly generating \(n\)-dimensional worlds that can represent the solution spaces for a wide range of SOPs.
That said, we do not predefine the optimization problem. Instead, the task naturally emerges as finding the maximum (or other extrema) in the generated \(n\)-dimensional world. This setup allows us to define flexible problems while preserving the integrity of the test environment.

\begin{figure}[!t]
\begin{center}
\centerline{
\includegraphics[width=\columnwidth]{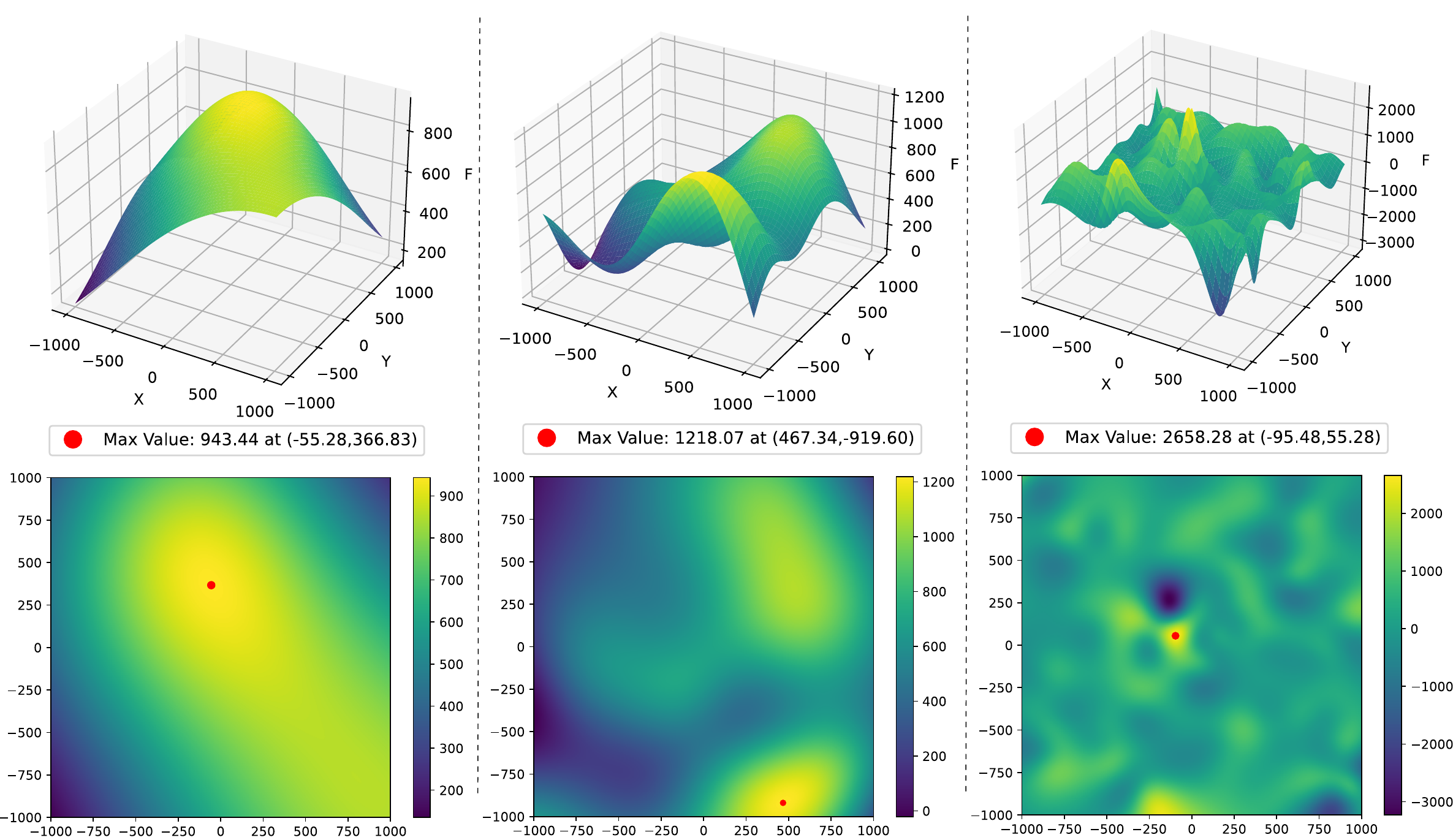}
}
\caption{Samples of generated 3-D worlds for different SOP complexities: very simple (left), simple (middle), and medium (right). The top plots display the 3-D graphs of the worlds, while the bottom plots show their 2-D heatmap versions.}
\label{fig:worlds}
\end{center}
\vskip -0.2in
\end{figure}

\textbf{Benefits and Advantages:} This approach ensures that neither the optimization problem nor its solution was exposed to the LLM during training. By doing so, we can mimic a real-world scenario where an optimization expert is asked to tackle a newly faced optimization problem using any techniques or strategies they prefer. Utilizing this approach brings some advantages. It offers generative flexibility, allowing the \(n\)-dimensional world to represent an infinite variety of optimization problems, from simple to highly complex ones. By abstracting the problem into a generated world, it ensures unbiased evaluation, reducing contamination from known problem-solution pairs and providing a more acceptable measure of the LLM's capability. Our world generator, WorldGen, enables the creation of increasingly complex worlds that test the limits of learning agents and provides a platform for benchmarking them under controlled yet dynamic conditions. 
Figure~\ref{fig:worlds} shows samples of generated 3-D worlds with different complexity levels.

\subsection{LLM and Accessing the World}
\begin{figure}[!t]
\begin{center}
\centerline{
\includegraphics[width=.8\columnwidth]{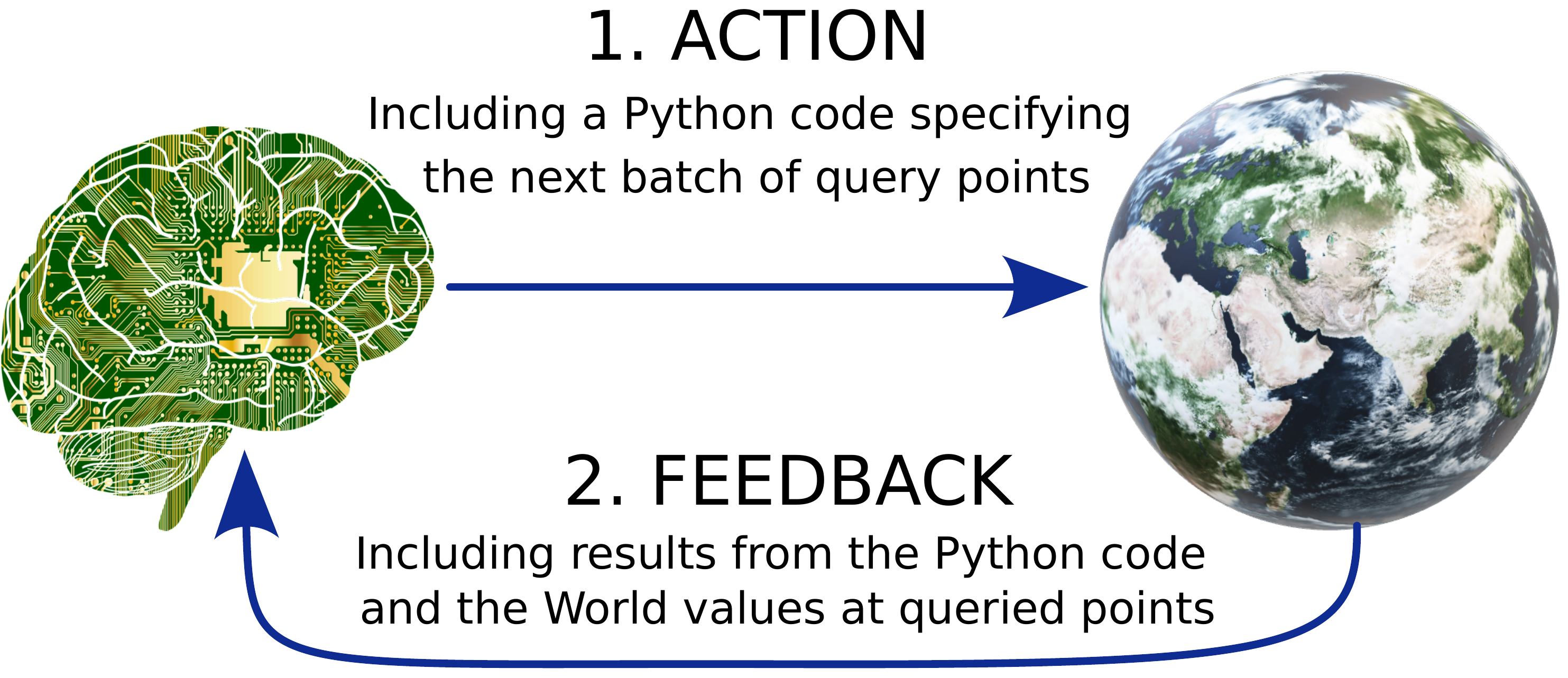}
}
\caption{Interaction loop between an LLM agent and the World}
\label{fig:interactive_cycle}
\end{center}
\vskip -0.4in
\end{figure}

\textbf{An Interactive Cycle:}
To enable the LLM agent to perform its task effectively, we provide it with access to the generated world through an interactive cycle as shown in Figure~\ref{fig:interactive_cycle}. This ensures a dynamic sequential process where the agent iteratively learns and refines its approach based on the information it gathers. 
Simply put, in each iteration, the LLM agent is allowed to interact with the world by selecting a batch of interested points where each point is a vector, $v_i$ of size $n-1$. Then, the world responds by revealing the corresponding values of $f(v_i)$ to the LLM agent. This will end one iteration/round of the interaction. In the next round, the LLM utilizes this feedback to determine the next set of points to query. 

\textbf{Supporting Coding \& Providing Flexibility:}
Due to the complexity of SOPs, the LLM agent is permitted to provide a Python code as part of its response and therefore utilize any library it deems necessary for solving the problem. This freedom ensures that the agent can employ a diverse range of tools and techniques to explore and analyze the generated world. As part of its role, the World is responsible for executing the Python code generated by the LLM agent. Once the code is executed, the World provides the results back to the agent as part of its feedback, enabling it to adapt and refine its strategy in subsequent iterations. In case of errors that may arise during execution, the World is responsible to provide the details of the errors and return meaningful feedback. Moreover, the LLM agent is not constrained by a fixed set of queries or techniques. Instead, it has the freedom to decide how to approach the problem, including leveraging mathematical models, heuristics, or machine learning techniques. This flexibility mimics the conditions under which human experts operate when solving optimization problems.
The interaction between the LLM and the World creates a real-time feedback loop. The LLM continually refines its understanding of the world based on the revealed data, while the World executes the agent’s strategies and provides results.

\subsection{Notion of Efficiency}
\textbf{To Solve or to Efficiently Solve:}
Exhaustively searching through all possible solutions, or brute force, is always a solution to any optimization problem; however, it represents the most inefficient approach. Thus, merely solving an optimization problem is not the primary goal; solving it efficiently is what truly matters. To formalize this, we require the notion of an efficient solution. But how can we define efficiency in a meaningful and practical way here? 

\textbf{The Expert Solution:}
To address this question, we designed a baseline referred to as the Expert Solution. This baseline serves as a reference point for assessing the efficiency of the LLM agent's performance. The Expert Solution is crafted using different optimization techniques, including a combination of Monte Carlo search methods, Bayesian optimization, and Active Learning strategies. 
An important aspect of the Expert Solution is the introduction of a \soheili{query budget}. This budget represents the number of queries required by the Expert Solution to reliably solve the optimization problem. It provides an upper bound on the number of interactions with the environment that are necessary to achieve a solution. That said, alongside the optimization problem, the LLM agent is provided with the query budget and instructed to not only solve the optimization problem but also do so within the given query budget.

\textbf{Incentivizing Efficiency:}
This setup ensures that the agent is incentivized to prioritize efficiency. It must strategize its queries, balancing exploration and exploitation to maximize the information gained from each interaction. By enforcing a query budget, we can objectively evaluate LLM's efficiency and effectiveness in solving the problem. Ultimately, the notion of efficient solutions pushes the LLM agent beyond simple problem-solving, encouraging it to adopt creative and resource-conscious strategies that align with real-world optimization challenges.

\subsection{Evaluating LLMs Performance in SOPs}
\label{sec:part_2_evaluations}

\textbf{The Setup:}
To evaluate the performance of the LLM agent, we follow a structured approach based on repeated experiments. We begin by generating worlds, characterized by a complexity index. In particular, to simplify experiments, visualizations, and keep overall token usage manageable, we focus on 3-D worlds and three levels of complexity: very simple (L0), simple (L1), and medium (L2). Next, we apply the Expert Solution to solve the corresponding SOPs associated with the generated world. As a result, we find a query budget required to achieve this reliably. Then, LLM agent is asked to solve the problem constrained by the query budget.
We repeat this process 100 times and measure the success rate of the LLM agent—defined as the proportion of trials where the agent successfully finds the optimal solution within the given query budget\footnote{We apply a relaxation here, treating any value found within 5\% of the optimum as the optimum point}. This success rate provides a quantitative measure of the agent's effectiveness and efficiency. In these experiments, we utilize the GPT-4-32k model as a capable baseline model. 

\textbf{The Default Scheme:}
Without involving any prompting techniques, the success rates become very low (close to 0\%). Therefore, we borrowed ideas from few-shot learning~\cite{brown2020fewshot}, Chain of Thought (CoT)~\cite{wei2022cot}, and added other techniques such as proper role assignment~\cite{karpathy2023stateofgpt} to improve the performance of the LLM agent. We name the resulting scheme \textit{LLM$^+$} and treat it as our default scheme from now on.
In particular, the prompt given to the LLM agent includes: [Role Assignment], followed by [Problem Definition \& Examples], [General Helpful Notes], and [Required Response Format]. The [Required Response Format] itself consists of [Plain Description of Current Strategy], [Python Code Implementation of the Strategy], and [Maximum Value Found So Far] fields (check Appendix~\ref{sec:app:prompts} for more details).

\textbf{Results:}

Table 1 summarizes the results of LLM$^+$ in various scenarios. These results optimistically suggest that LLMs understand SOP settings and are familiar with optimization techniques. As expected, the success rate of LLM$^+$ depends on the complexity of the underlying world. However, it is somewhat surprising that even in relatively simple scenarios (L1), LLM$^+$ does not achieve a high success rate. In straightforward settings where the world representing the SOP has only one global optimum with no other local optimum points, LLM$^+$ effectively utilizes general optimization techniques such as gradient ascent. But, when the world exhibits some complexity, with a few local optimum points in non-trivial parts of the space—common in real-world scenarios—LLMs struggle to strategize properly and find the global optimum (check Appendix~\ref{sec:app:llm+} for more details).

\begin{table}[t]
\caption{The success rates of LLM$^+$ (with GPT-4-32k base model) in different worlds and complexity levels}
\label{table:primary_results}
\begin{center}
\begin{small}
\begin{sc}
\begin{tabular}{l|c}
\toprule
World Complexity & Success Rate $\uparrow$\\
\hline
L0 (Very Simple) & 100\% \\
L1 (Simple) & 36\% \\
L2 (Medium) & 4\% \\

\bottomrule
\end{tabular}
\end{sc}
\end{small}
\end{center}
\vskip -0.3in
\end{table}

\subsection{A Dialectical Perspective to Enhance LLMs in SOPs}
\label{sec:part_3_ace}

Motivated by the unsatisfactory performance of vanilla LLMs in non-straightforward SOPs, we aim to address a natural follow-up question: Can we enhance the performance of LLMs without relying on retraining, fine-tuning, or post-training modifications? 

\textbf{The Core Idea:}
To that end, we propose a framework inspired by the well-established Hegelian Dialectics, offering a formal structure for improving LLM performance through dynamic dialectical reasoning processes.
Using the terminology of Hegelian Dialectics, we can conceptualize a general LLM agent as a Thesis Generator entity—a block that observes a given problem and generates a corresponding solution or response to it\footnote{Note that this remains valid even within Minsky's multi-agent system, where the main problem is divided into smaller pieces, each delegated to different agents communicating with one another.}. However, inspired by Hegel's framework, we suggest that confining to a single Thesis Generator block is not an appropriate model for solving problems or generating ideas. Instead, a more robust model should include additional components: an Antithesis Generator and a Synthesis Block. Together, these three components, as Hegel formally describes~\cite{hegel_phenomenology_spirit,hegel_science_logic}, form a dynamic reasoning cycle, enabling the system to iteratively refine and improve its outputs.

\begin{figure}[!t]
\begin{center}
\centerline{
\includegraphics[width=\columnwidth]{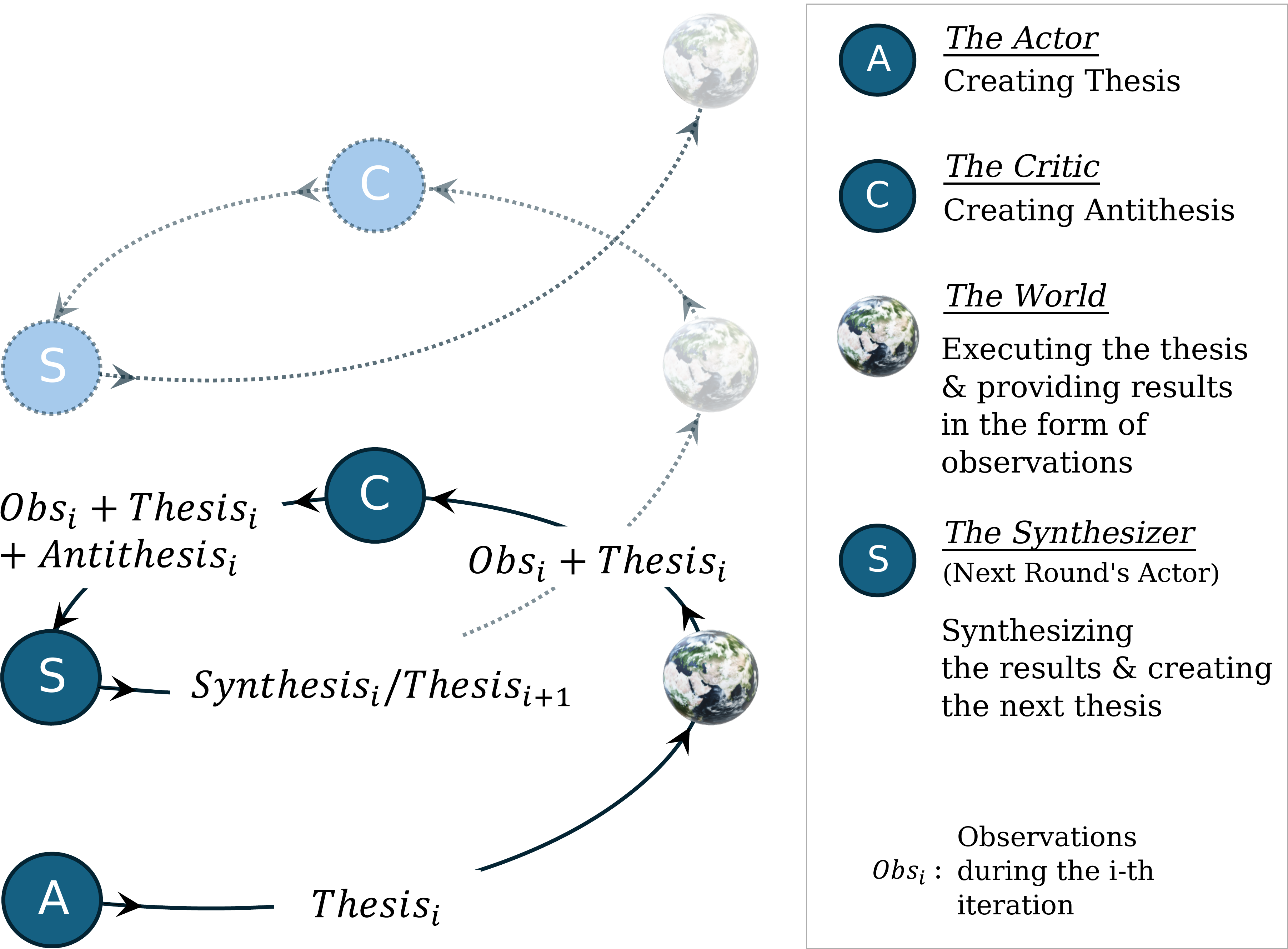}
}
\caption{ACE and the spiral of thoughts}
\label{fig:ace}
\end{center}
\vskip -0.4in
\end{figure}

\textbf{ACE:}
The Antithesis Generator plays a critical role by challenging the solutions produced by the Thesis Generator. It identifies potential flaws, contradictions, or alternative perspectives that may have been overlooked. This counterbalance forces the system to evaluate its assumptions critically and consider a broader range of possibilities. The Synthesis Block then reconciles the Thesis and Antithesis, combining their insights to produce a more refined and coherent solution. This iterative interplay between Thesis, Antithesis, and Synthesis ensures that the system continuously evolves its understanding and response, ultimately arriving at a more suitable outcome.
This dialectical structure led us to introduce our solution, \textit{ACE}\footnote{ACE stands for \textbf{A}ct, \textbf{C}ritique, and \textbf{E}volve}, embodying three components: (1) Actor, (2) Critic, and (3) Synthesizer. The relationship between these components is shown in Fig.~\ref{fig:ace}.
The iterative cycle of solving a problem starts with the Actor creating an initial thesis. This thesis is then implemented and executed in the world, and the corresponding outcomes and results (called observations) are gathered. Next, the Critic examines the initial thesis and the corresponding observations to generate an antithesis. The thesis, antithesis, and corresponding observations are then fed into the Synthesizer. The Synthesizer creates an evolved thesis, completing an iteration/round. The cycle continues by treating the evolved thesis as the next initial thesis in the cycle.
As we later show in section~\ref{sec:eval}, ACE significantly improves the performance of LLMs with no modification to their architecture and no extra post-training or fine-tuning. By embedding a dialectical reasoning process into the system, ACE creates a more adaptive process that is better equipped to handle complex and nuanced sequential optimization tasks (Appendix~\ref{sec:app:dialectical_sample} provides samples and details of ACE's dialectical process).

\section{Evaluation}
\label{sec:eval}
\begin{table*}[t]
\caption{Success rates (\%) of different schemes in 3-D worlds with L1 and L2 complexity levels using various base LLMs}
\label{table:main}
\begin{center}
\begin{small}
\begin{sc}
\resizebox{\textwidth}{!}{
\begin{tabular}{l|ccc|ccc}
\toprule
Scheme & \multicolumn{6}{c}{Success Rate $\uparrow$}  \\
\hline
& \multicolumn{3}{c|}{L1} & \multicolumn{3}{c}{L2}\\
\cline{2-7}
\multirow{2}{*}{} & GPT-4-32k & Llama-3-70B-Inst & GPT-3.5-Turbo & GPT-4-32k & Llama-3-70B-Inst & GPT-3.5-Turbo \\
\cline{2-7}
LLM$^+$ (Default)       & 36\% & 16\% & 7\% & 4\% & \textbf{3\%} & 1\%\\ 
Self-Reflection      & 30\% & 16\% & 8\% & 7\% & \textbf{3\%} & \textbf{3\%}\\
Debate               & 39\% & 16\% & 7\% & 7\% & 2\% & 1\%\\
Majority             & 38\% & 27\% & 10\% & 7\% & \textbf{3\%} & 2\%\\
ACE                  & \textbf{88\%} & \textbf{29\%} & \textbf{22\%} & \textbf{9\%} & 2\% & \textbf{3\%}\\
\bottomrule
\end{tabular}
}
\end{sc}
\end{small}
\end{center}
\vskip -0.2in
\end{table*}

\subsection{Overall Results}
\textbf{Settings:}
To put the performance improvements of ACE in proper context, we implemented several recent related proposals including Self-Reflection~\cite{madaan2024selfreflection}, Majority Vote~\cite{wang2022majority,lewkowycz2022majority2}, and Debate~\cite{du2023debate} and compared them with ACE. To have a fair comparison, for the Majority and Debate schemes we set the total number of agents to three and two agents, respectively, to roughly match the token usage of ACE. Later, in section~\ref{sec:eval:deepdive}, we perform more evaluations with higher number of agents for them.
We accompany all these schemes with the additional prompting techniques appeared in LLM$^+$ to have a fair comparison (check Appendix~\ref{sec:app:prompts} for more details). As for the LLMs, we use 3 different models: GPT-4-32K~\cite{gpt4}, Llama-3-70B-Instruct~\cite{dubey2024llama3}, and GPT-3.5-Turbo~\cite{openai2025gpt35}. As before, we repeat evaluations 100 times and report the success rate of different schemes.

\textbf{Results:} Table~\ref{table:main} presents the findings. A comparison of the general performance of LLMs in L1 and L2 complexity levels reveals that LLMs struggle to solve SOPs as problem complexity increases. This aligns with the objective of WorldGen, which is designed to generate progressively complex SOPs to rigorously evaluate model performance. The results also demonstrate that ACE significantly enhances performance, boosting the capabilities of base language models across various scenarios. For instance, ACE achieves a remarkable success rate of 88\% with GPT-4-32K in L1 complexity, more than doubling the next best approach.
When compared to alternative methods, ACE emerges as the superior solution, showcasing its effectiveness in enhancing LLM performance. However, ACE is not without limitations. In scenarios where the base model’s inherent capabilities are limited, such as in L2 complexity with GPT-3.5-Turbo, performance gains are modest. This is because ACE operates by treating LLM as a black-box, relying on the existing abilities of the LLM without retraining or modifying its weights. Consequently, ACE cannot compensate for a model’s lack of foundational knowledge. For example, a model achieving a 0\% success rate indicates it lacks the necessary baseline understanding to solve the problem, rendering ACE's dialectical method—rooted in the same model—ineffective. That explains why ACE's effectiveness varies with the capabilities of the underlying LLM\footnote{A sample dialectical process in ACE is shown in Appendix~\ref{sec:app:dialectical_sample}}.

\subsection{Deep Dive}
\label{sec:eval:deepdive}
\textbf{Cost Comparison:}
We evaluate the cost of using ACE by analyzing the total number of tokens consumed and comparing it with other approaches. Table~\ref{table:cost} provides the average total token usage for one run of the experiment for various schemes, and their corresponding normalized values (to the default single-agent scheme, LLM$^+$) using GPT-3.5-Turbo as the base model. On average, ACE consumes $2.27\times$ the total tokens compared with LLM$^+$; however, it remains more efficient compared to multi-agent schemes like Debate and Majority, which require higher token usage.

\begin{table}[t]
\caption{Averaged total consumed tokens of different schemes (with GPT-3.5-Turbo as base) in one instance of an experiment}
\label{table:cost}
\begin{center}
\begin{small}
\begin{sc}
\resizebox{0.9\columnwidth}{!}{%
\begin{tabular}{lcc}
\toprule
Scheme & Normalized $\downarrow$ & Total Tokens$\downarrow$ \\
\midrule
LLM$^+$            & 1  &  6912 \\
Self-Reflection & 1.24$\times$   &  8581 \\
ACE             & 2.27$\times$   &  16211 \\
Debate          & 2.66$\times$   &  18396 \\
Majority        & 3.29$\times$   &  22733 \\
\bottomrule
\end{tabular}
}
\end{sc}
\end{small}
\end{center}
\vskip -0.2in
\end{table}

\textbf{ACE vs. Multi-Agent Schemes with More Agents:} 
A natural assumption might be that increasing the number of agents in schemes like Debate and Majority would enhance performance. So, we conducted additional experiments, scaling the number of agents to seven (referred to as Debate$^\star$ and Majority$^\star$) and comparing their performance and token costs with ACE. These experiments were carried out using GPT-4-32K and GPT-3.5-Turbo as base models in the L1 3-D world sample. The results are summarized in Table~\ref{table:7agents}.
As shown in Table~\ref{table:7agents}, increasing the number of agents in these schemes results in higher costs without corresponding performance improvements. In fact, the performance of Debate$^\star$ and Majority$^\star$ degrades when compared to their counterparts with fewer agents. Schemes like Debate$^\star$, which require exchanging responses among all agents in each round, can experience exponential growth in token consumption as the number of agents increases. This inefficiency is particularly evident in SOP settings, where sequential decision-making processes require multiple rounds of interaction, as illustrated by the token cost data in Table~\ref{table:7agents}.
The key takeaway is that simply increasing the number of agents in tasks involving sequential decision-making, such as those in SOP settings, does not necessarily yield better results. Instead, it often introduces inefficiencies and performance degradation in solutions like Debate$^\star$ and Majority$^\star$.

\begin{table}[h]
\caption{Success rates (\%) and averaged total token cost of Debate$^\star$ and Majority$^\star$ (using 7 agents) and ACE, normalized to tokens used by LLM$^+$ (with the same base LLM) in L1 3-D world scanerio}
\label{table:7agents}
\begin{center}
\begin{small}
\begin{sc}
\resizebox{0.98\columnwidth}{!}{
\begin{tabular}{l|cc|cc}
\toprule
Scheme & \multicolumn{2}{c|}{GPT-4-32k} & \multicolumn{2}{c}{GPT-3.5-Turbo} \\
\hline
 & Success $\uparrow$ & Cost $\downarrow$ & Success $\uparrow$ & Cost $\downarrow$  \\
\hline
Debate$^\star$     & 32\% & 32.07$\times$ & 3\% &  11.34$\times$\\
Majority$^\star$   & 27\% & 6.34$\times$ & 6\% & 6.83$\times$ \\
ACE         & \textbf{88}\% & \textbf{3.64$\times$} & \textbf{22\%} & \textbf{2.27$\times$} \\
\bottomrule
\end{tabular}
}
\end{sc}
\end{small}
\end{center}
\vskip -0.1in
\end{table}

\textbf{ACE in Static Settings:}
Our primary domain and targeted setting (SOPs) possess an important characteristic: the ability of the World to provide feedback during task execution. This feedback plays a pivotal role in ACE’s architecture, as the generation of Antithesis relies not only on the Thesis but also on the feedback provided by the World after executing the Thesis. This feedback serves as a ground truth, guiding the generation of a more refined Antithesis. 
To explore how ACE performs in scenarios without access to such feedback, we use a modified version of ACE, referred to as ACE$^\ast$, which operates in a general static task where real-time feedback from the World is unavailable. So here, we utilize the MMLU~\cite{mmlu} benchmark, focusing on multiple-choice question answering. For our experiments, we select three datasets: (1) a set of 100 randomly chosen questions (Set \#1), (2) a college-level Physics test (Set \#2), and (3) a high-school-level Statistics test (Set \#3). 
Table~\ref{table:mmlu} summarizes the results. While ACE$^\ast$ outperforms other schemes like Debate and Majority in Sets \#1 and \#2, the performance gap is less pronounced than in SOP context where feedback from the World is available. Additionally, ACE$^\ast$ struggles with Set \#3, showing lower performance levels compared to other benchmarks. These results can highlight the critical importance of World feedback in effective Antithesis generation.

\begin{table}[t]
\caption{Averaged Score of Different Schemes (with GPT-4-32K as base model) in MMLU benchmarks}
\label{table:mmlu}
\begin{center}
\begin{small}
\begin{sc}
\resizebox{\columnwidth}{!}{%
\begin{tabular}{lccc}
\toprule
Scheme & MMLU (set \#1) & MMLU (set \#2) & MMLU (set \#3)\\
\midrule
Default         & 87.1 $\pm$ 0.74 & 80.2 $\pm$ 2.1 & 179.6 $\pm$ 2.67\\
Majority        & 87.3 $\pm$ 0.95 & 81.1 $\pm$ 1.1 & \textbf{180.7} $\pm$ 1.7\\
Debate          & 87.3 $\pm$ 1.64 & 69.6 $\pm$ 3.1 & 160.0 $\pm$ 3.23\\
ACE$^\ast$             & \textbf{88.4} $\pm$ 2.32 & \textbf{85.0} $\pm$ 2.21 & 178.1 $\pm$ 3.07\\
\bottomrule
\end{tabular}
}
\end{sc}
\end{small}
\end{center}
\vskip -0.3in
\end{table}

\section{Limitations \& a Brief Discussion}
\textbf{WorldGen's Limitation:}
WorldGen effectively generates worlds with adjustable complexity for testing LLMs in SOPs, but relies on manually designed Expert solutions to solve the SOP. This dependence on human expertise for robust baselines can be time-intensive and limit the automation potential of the approach. We leave addressing the fully automated objective to future work.

\textbf{Limitations of ACE:}
ACE's dialectical framework has demonstrated great performance in our main targeted domain, SOPs, but its effectiveness in other domains, particularly those lacking real-time feedback, remains an open question. In static question-answering or static tasks without iterative refinement, the benefits of ACE may be limited, warranting further evaluation. Additionally, by treating LLMs as black boxes, ACE's performance is inherently bound by the capabilities of the underlying model. Moreover, while its token consumption is lower than multi-agent schemes like Debate or Majority, ACE incurs a slight overhead compared to single-agent approaches. This trade-off is minor in complex SOP tasks but could pose challenges in resource-constrained scenarios.

\textbf{On the Potential of ACE:} 
The potential of ACE, rooted in its Hegelian dialectical framework, extend beyond solving SOPs. Its dialectical approach, mirroring human-like problem-solving processes, fosters solutions that are not only accurate but also deeply contextual and well-reasoned. Furthermore, Hegelian philosophy provides a foundation to explain the effectiveness of other prompt engineering techniques, such as self-reflection, by framing them within a structured dialectical process. This perspective can deepen our understanding of existing methods and their mechanisms.
Additionally, the Hegelian-inspired framework offers a powerful structure for \textit{generating synthetic data}. Its iterative nature facilitates the creation of diverse, high-quality datasets that reflect a broad range of perspectives and solutions, making them invaluable for training and fine-tuning LLMs to tackle complex and nuanced tasks effectively.

\textbf{LLM$^+$ Could Have Been Better!}
A fair criticism might be that LLMs might perform better in solving SOPs with improved prompt engineering. We are not claiming that LLM$^+$ represents the optimal default scheme; rather, we argue that it serves as a robust baseline. Even with carefully designed prompts, LLMs' performance in this setting remains limited, highlighting the need for approaches like ACE to unlock their full potential and deliver superior performance.

\textbf{What If the Next LLM Becomes Very Capable?} 
A more capable LLM makes ACE even more useful, not less. As demonstrated in Section~\ref{sec:eval}, a better base model serves as a stronger foundation, enabling even greater performance improvements. In essence, ACE with its dialectical base is designed to complement and amplify the capabilities of any LLM, regardless of its initial proficiency in SOP context. By leveraging ACE, we can transform an already impressive LLM into an extraordinary one, pushing what is possible and unlocking new levels of performance in this domain.

\section{Final Note}
Our exploration into the capabilities of LLMs in tackling sequential optimization problems has revealed both their potential and their current limitations. Through the development and use of WorldGen, we have shown that while LLMs exhibit impressive abilities, they still face challenges with even relatively simple SOPs. These findings have led us to propose a novel approach inspired by philosophical reasoning frameworks, aiming to enhance LLM performance in innovative yet easy to reason about ways.
We believe that this work can open new avenues, encouraging the integration of philosophical reasoning frameworks into AI systems. By fostering a deeper understanding and application of these frameworks, we can pave the way for more robust and intelligent systems. We hope our efforts inspire others to explore these interdisciplinary approaches, ultimately contributing to the advancement of LLMs and its applications across diverse fields.



\bibliography{ref}
\bibliographystyle{icml2025}


\usetikzlibrary{fit, positioning, backgrounds}

\newpage
\appendix
\onecolumn
\section{More on the Evaluations and the Prompt Templates Used}
\label{sec:app:prompts}
\textbf{Prompt Templates:}
Figure~\ref{fig:prompts:llm+} shows the main template used for the LLM$^+$ scheme. We use the same template for the main agent of other schemes compared in this paper, including ACE's Actor. Additionally, Figure~\ref{fig:prompts:critic} and~\ref{fig:prompts:critic:transitional} demonstrate the initial and transitional prompts used for ACE's Critic, respectively. The task of the Synthesizer, the Actor of the previous and next steps, will be identified through a transitional prompt, as shown in Figure~\ref{fig:prompts:synth}.

\textbf{Majority Scheme:}
To implement the Majority scheme and automate the solution, we use another agent called the poll worker. The poll worker checks different agents' responses and identifies the Majority response, which is the one with the highest consensus. Figure~\ref{fig:prompts:pollworker} shows the prompt template used for the poll worker. 
Unlike taking the majority vote after every agent completes the task in general scenarios, in our sequential decision-making problems, we need to take the majority vote in every round. Therefore, the poll worker processes the agents' responses at each round of interaction with the World, identifies the response with the majority consensus at each round so that the World can execute it and provide the feedback.

\section{A Couple of Samples for LLM$^+$ in Action} 
\label{sec:app:llm+}
Figures~\ref{fig:llm+:sample1} and~\ref{fig:llm+:sample2} demonstrate two samples of strategies used by LLM$^+$ in separate runs, utilizing the GPT-4-32K base model in an L1 world. In the first run (Figure~\ref{fig:llm+:sample1}), the agent begins with a broad grid search strategy, aiming to cover the entire search space and identify regions with high values. This initial phase can form a foundation for subsequent searches. However, the agent does not adapt its strategy in the following rounds, remaining static and failing to find the optimum point by the end of the 16 rounds.

In the second example (Figure~\ref{fig:llm+:sample2}), the agent attempts to adapt its strategy based on the feedback it receives from the World. Initially, the agent starts with a coarse exploration to identify promising regions. It then refines the search around one promising area identified in the initial phase and continues with further refinements. However, it still cannot find the optimum point. This time, the issue lies in the agent spending a significant number of queries exploring around the local maximum found. The search strategy was not sufficiently adaptive to balance between local exploitation and global exploration. Nevertheless, the fact that it found a local maximum is noteworthy, as it demonstrates some degree of understanding about the notion of a maximum point in a region, the direction of increase or decrease of a sequence of observations, and their relation to the actual curve modeling the unknown World function, \( f \).

\section{A Sample of the Dialectical Process in ACE} 
\label{sec:app:dialectical_sample}
Figures~\ref{fig:ace:sample:part1} and~\ref{fig:ace:sample:part2} demonstrate a sample of a dialectical progress with ACE, utilizing the GPT-4-32K base model in an L1 world. The iterative process of ACE's dialectical method, which involves a cycle of thesis, antithesis, and synthesis allows for systematic refinement and improvement of strategies based on feedback and critique. 
One of the key advantages of ACE is its ability to adapt and improve through structured feedback. For instance, the initial grid search strategy (Thesis 1 in Figure~\ref{fig:ace:sample:part1}) provided a broad understanding of the search space. However, the corresponding antithesis highlighted the need for refinement in promising areas, the incorporation of adaptive techniques, and a balance between exploration and exploitation. This critical feedback led to a more refined and effective strategy (Synthesis 1 in Figure~\ref{fig:ace:sample:part2}), which combined grid search with adaptive methods like simulated annealing.

The dialectical method also ensured that the agent did not become overly focused on a single approach. By evaluating and adjusting strategies through a dialectal cycle, ACE attempted to balance the thorough exploration of high-value areas with the need to investigate less explored regions. This is evident in the transition from Thesis 2 to Synthesis 2 (Figure~\ref{fig:ace:sample:part2})), where the agent broadened its exploration based on feedback, expanding the use of simulated annealing to uncover potential peaks outside the heavily focused regions.

Moreover, the dialectical approach fostered a dynamic and flexible search process. The agent's ability to incorporate feedback and adjust its methods in real-time allowed for efficient use of queries and increased the chances of identifying the global maximum. This adaptability is crucial in complex search spaces where the landscape can vary significantly.

In short, the Hegelian Dialectics aspect of ACE offered a mechanism for continuous improvement. By leveraging structured feedback and iterative refinement, ACE enhanced the agent's ability to navigate search spaces effectively. The figures illustrate this process, showcasing how each cycle of thesis, antithesis, and synthesis leads to progressively better strategies and outcomes.

\begin{figure}[t!]
    \centering
    \begin{tikzpicture}
        \pgfdeclarelayer{background}
        \pgfsetlayers{background,main}

        \node[align=center, text=black, font=\large\bfseries] (title) {Main Initial Prompt Template for LLM$^+$ and Other Schemes};
 
        \node[fill=red!30, draw=SlateGray2, rounded corners, inner sep=5pt, text width=\textwidth, below=of title, yshift=0.2cm] (prompt1) {
            \small
            \textit{You are a great expert in the optimization topic and search algorithms.}
        };
        \node[above=0.25cm of prompt1.north west, anchor=west] {\textbf{Role Assignment}};

        \node[fill=green!30, draw=SlateGray2, rounded corners, inner sep=5pt, text width=\textwidth, below=of prompt1, yshift=0.2cm] (prompt2) {
            \small
            You are tasked with examining an unknown function \( f(x,y) \) \((-1000 \leq x,y \leq 1000)\). You need to interact with the function \( f(x,y) \) in order to locate the global maximum value. Here's how to do it:
            \begin{enumerate}
                \item \textit{Define Your Strategy}: Start with creating a solid strategy to explore the space and solve the problem.
                \item \textit{Choose a Point (x, y)}: Based on your strategy, select unique NEW points \((x_1, y_1), (x_2, y_2), \ldots\) to evaluate the function.
                \item \textit{Get Feedback and Adjust Your Strategy}: After I reveal the values of \((x_1, y_1, f_1), (x_2, y_2, f_2), \ldots\) at your chosen points, adjust your strategy based on this feedback.
                \item \textit{Repeat the Process}: Continue this process for up to \textit{$QueryBudget$} queries (in the form of \((x_i, y_i)\)) or until you are confident that you have found the global maximum.
            \end{enumerate}
            
            \textit{Note}: Finding a value in the range of \([0.95 \times (\text{Global Max}), \text{Global Max}]\) is equal to solving the problem.
        };
        \node[above=0.25cm of prompt2.north west, anchor=west] {\textbf{Problem Definition}};

        \node[fill=orange!30, draw=SlateGray2, rounded corners, inner sep=5pt, text width=\textwidth, below=of prompt2, yshift=0.2cm] (prompt3) {
            \small
            Here's how you should format your response:
            \begin{itemize}
                \item \textit{MY\_CURRENT\_STRATEGY}: $<$explain your chosen strategy here$>$
                \item \textit{MAX\_SEEN\_SO\_FAR}: $x,y, f(x,y)$ 
                \item \textit{NEXT}: $<$Python code snippet that generates the next coordinates and return a list of tuples [(xi, yi), ...]$>$
            \end{itemize}
            \{...\}          
        };
        \node[above=0.25cm of prompt3.north west, anchor=west] {\textbf{Response Format}};

        \node[fill=cyan!30, draw=SlateGray2, rounded corners, inner sep=5pt, text width=\textwidth, below=of prompt3, yshift=0.2cm] (prompt4) {
            \small
            Here are some rules that you must follow: \{...\}
        };
        \node[above=0.25cm of prompt4.north west, anchor=west] {\textbf{General Rules and Examples of Acceptable/Unacceptable Responses}};

        \node[fill=yellow!30, draw=SlateGray2, rounded corners, inner sep=5pt, text width=\textwidth, below=of prompt4, yshift=0.2cm] (prompt5) {
            \small
            \begin{itemize}
                \item The space is vast, and there will be several LOCAL maximums, so avoid choosing them as the answer. Make sure that you explore the space enough to ensure that your answer represents the GLOBAL maximum  
                \item Asking for a certain coordinates multiple times, consumes your available remaining query budget and reduces your chances of finding the global maximum. So, utilize the responses so far to select only unique coordinates.
                \item A very important point is that YOU SHOULD NOT BE HASTY. You should be patient and explore the space thoroughly. However, remember that you have only a maximum of \textit{$QueryBudget$} queries to solve the problem.
            \end{itemize}
            \{...\}
        };
        \node[above=0.2cm of prompt5.north west, anchor=west] {\textbf{General Hints}};

        \node[fill=magenta!30, draw=SlateGray2, rounded corners, inner sep=5pt, text width=\textwidth, below=of prompt5, yshift=0.2cm] (prompt6) {
            \small
            Here are examples of function $f$ where it has multiple local maxima and one global maximum \{...\}
        };
        \node[above=0.2cm of prompt6.north west, anchor=west] {\textbf{Examples of Function $f$}};

        \node[fill=lightgray!10, draw=SlateGray2, rounded corners, inner sep=5pt, text width=\textwidth, below=of prompt6, yshift=0.2cm] (prompt7) {
            \small 
            Let's start. Create an excellent and efficient strategy and choose your first batch of coordinates accordingly.
        };
        \node[above=0.2cm of prompt7.north west, anchor=west] {\textbf{Start Command}};

        \begin{pgfonlayer}{background}
            \node[fit=(title)(prompt1)(prompt2)(prompt3)(prompt4)(prompt5)(prompt6)(prompt7), fill=blue!10, rounded corners, inner sep=10pt] {};
        \end{pgfonlayer}
    \end{tikzpicture}
    \vspace{-2em}
    \caption{Main initial prompt template used for LLM$^+$ and other schemes}
    \label{fig:prompts:llm+}
\end{figure}
\begin{figure}[t!]
    \centering
    \begin{tikzpicture}
        \pgfdeclarelayer{background}
        \pgfsetlayers{background,main}

        \node[align=center, text=black, font=\large\bfseries] (title) {Critic's Initial Base Prompt Template};

        \node[fill=red!30, draw=SlateGray2, rounded corners, inner sep=5pt, text width=\textwidth, below=of title, yshift=0.2cm] (prompt1) {
            \small
            \textit{You are a great expert in the optimization topic and search algorithms and will assist others in solving optimization problems.}
        };
        \node[above=0.25cm of prompt1.north west, anchor=west] {\textbf{Role Assignment}};

        \node[fill=green!30, draw=SlateGray2, rounded corners, inner sep=5pt, text width=\textwidth, below=of prompt1, yshift=0.2cm] (prompt2) {
            \small 
            Your task is to provide guidance, suggestions, and assistance to a very smart AI agent for solving an optimization problem.
            
            The agent is to interact with an unknown function \( f(x,y) \) \((-1000 \leq x, y \leq 1000)\) with the objective of identifying the global maximum value. Here is the procedure the agent will adhere to:
            
            \begin{enumerate}
                \item \textit{Strategy Development}: The agent will begin by devising a comprehensive strategy to explore the space and tackle the problem.
                \item \textit{Point Selection (x, y)}: The agent will choose unique NEW points \([(x_1, y_1), (x_2, y_2), \ldots]\) for function evaluation, based on its strategy.
                \item \textit{Feedback Collection and Strategy Enhancement}: Once the values of \([(x_1, y_1, f_1), (x_2, y_2, f_2), \ldots]\) at the agent's selected points are revealed, the agent can refine its strategy using this feedback.
                \item \textit{Process Persistence}: The agent will continue this procedure for up to \textit{$QueryBudget$} queries (in the form of \((x_i, y_i)\)) or until it is confident that the global maximum has been identified.
            \end{enumerate}
            
            The agent should present its findings in the following way: \{...\}

            The agent will comply with the following rules: \{...\}

            Given this problem statement, your duty is to ensure that the agent identifies the global maximum value.
        };
        \node[above=0.25cm of prompt2.north west, anchor=west] {\textbf{Problem Definition}};

        \node[fill=yellow!30, draw=SlateGray2, rounded corners, inner sep=5pt, text width=\textwidth, below=of prompt2, yshift=0.2cm] (prompt3) {
            \small
            To achieve your goal, please follow these guidelines:
            \begin{enumerate}
                \item After each step, you can critique the agent's chosen coordinates or its strategy. It is essential that you offer constructive criticism to improve its next moves.
                \item It's important to remember that the space is vast, and there may be several LOCAL maximums, so you must help the agent avoid mistaking local maximum values for the answer \{...\}
                \item You can offer suggestions and brainstorming to assist the agent in its task. Remember that the agent is very smart, so do not describe what the agent has already chosen or done! Limit your responses to constructive criticism.
                \item Note: Finding a value in the range of \([0.95 \times (\text{Global Max}), \text{Global Max}]\) is equal to solving the problem, so discourage the agent to spend time on finding values that have small differences.
                \item After every iteration, list the potential issues with the agent's strategy and decision so far.
                \item Ensure that your responses are concise and to the point. Do not provide unnecessarily long responses.
            \end{enumerate}
            \{...\}
        };
        \node[above=0.25cm of prompt3.north west, anchor=west] {\textbf{General Guidelines}};

        \node[fill=cyan!30, draw=SlateGray2, rounded corners, inner sep=5pt, text width=\textwidth, below=of prompt3, yshift=0.2cm] (prompt4) {
            \small
            Here is the Agent's response: $<Thesis_1>$ and the corresponding results: $<Observations_1>$
        };
        \node[above=0.25cm of prompt4.north west, anchor=west] {\textbf{Agent's Response \& the Corresponding Results}};




        \node[fill=lightgray!10, draw=SlateGray2, rounded corners, inner sep=5pt, text width=\textwidth, below=of prompt4, yshift=0.2cm] (prompt5) {
            \small
            Now, given all the info, review agent's response, make your criticism and suggestions, and detect potential issues ...
        };
        \node[above=0.2cm of prompt5.north west, anchor=west] {\textbf{Start Command}};

        \begin{pgfonlayer}{background}
            \node[fit=(title)(prompt1)(prompt2)(prompt3)(prompt4)(prompt5), fill=blue!10, rounded corners, inner sep=10pt] {};
        \end{pgfonlayer}
    \end{tikzpicture}
    \vspace{-2em}
    \caption{Initial prompt template used for the Critic in ACE}
    \label{fig:prompts:critic}
\end{figure}

\begin{figure}[t!]
    \centering
    \begin{tikzpicture}
        \pgfdeclarelayer{background}
        \pgfsetlayers{background,main}

        \node[align=center, text=black, font=\large\bfseries] (title) {Critic's Transitional Prompt Template};

        \node[fill=cyan!30, draw=SlateGray2, rounded corners, inner sep=5pt, text width=\textwidth, below=of title, yshift=0.2cm] (prompt4) {
            \small
            Here is the Agent's response: $<Thesis_i>$ and the corresponding results: $<Observations_i>$
        };
        \node[above=0.25cm of prompt4.north west, anchor=west] {\textbf{Agent's Response \& the Corresponding Results}};




        \node[fill=lightgray!10, draw=SlateGray2, rounded corners, inner sep=5pt, text width=\textwidth, below=of prompt4, yshift=0.2cm] (prompt5) {
            \small
            Now, given all the info, review agent's response, make your criticism and suggestions, and detect potential issues ...
        };
        \node[above=0.2cm of prompt5.north west, anchor=west] {\textbf{Start Command}};

        \begin{pgfonlayer}{background}
            \node[fit=(title)(prompt4)(prompt5), fill=blue!10, rounded corners, inner sep=10pt] {};
        \end{pgfonlayer}
    \end{tikzpicture}
    \vspace{-2em}
    \caption{Transitional prompt template used for the Critic in ACE}
    \label{fig:prompts:critic:transitional}
\end{figure}
\begin{figure}[t]
    \centering
    \begin{tikzpicture}
        \pgfdeclarelayer{background}
        \pgfsetlayers{background,main}
        \node[align=center, text=black, font=\large\bfseries] (title) {Synthesizer's Prompt Template};
        \node[fill=green!30, draw=SlateGray2, rounded corners, inner sep=5pt, text width=\textwidth, below=of title, yshift=0.2cm] (prompt1) {
            \small 
            The corresponding results are: $<Observations_i>$
            
            \vspace{1em}
            
            To help you on your task, we provide you (the Agent/Actor) with the response from a reviewer who is observing your attempts: 
            
            \vspace{1em}
            
            $<Antithesis_i>$
            
            \vspace{1em}
            
            Given the suggestions and comments provided, improve your strategy and continue.
        };
        \node[above=0.25cm of prompt1.north west, anchor=west] {\textbf{Synthesize Command}};

        \begin{pgfonlayer}{background}
            \node[fit=(title)(prompt1), fill=blue!10, rounded corners, inner sep=10pt] {};
        \end{pgfonlayer}
    \end{tikzpicture}
    \vspace{-2em}
    \caption{The prompt template used for the Synthesizer in ACE. Note that Synthesizer is the Actor of the previous round, so it already has access to the $Thesis_i$. This provides an  efficient handling of the context and token usage.}
    \label{fig:prompts:synth}
\end{figure}
\begin{figure}[ht!]
    \centering
    \begin{tikzpicture}
        \pgfdeclarelayer{background}
        \pgfsetlayers{background,main}

        \node[align=center, text=black, font=\large\bfseries] (title) {Poll Worker Prompt Template};

        \node[fill=red!30, draw=SlateGray2, rounded corners, inner sep=5pt, text width=\textwidth, below=of title, yshift=0.2cm] (prompt1) {
            \small
            \textit{You are a great assistant with a strong background in AI and optimization problems.}
        };
        \node[above=0.25cm of prompt1.north west, anchor=west] {\textbf{Role Assignment}};

        \node[fill=green!30, draw=SlateGray2, rounded corners, inner sep=5pt, text width=\textwidth, below=of prompt1, yshift=0.2cm] (prompt2) {
            \small 
            You are assigned to work as a poll worker to analyze responses from multiple agents to a given problem. Each agent’s response may include long sentences describing their strategy to solve the problem, code snippets, or other relevant information.
            
            Your task is to identify the agent whose response is the most frequently specified among all agents. If there is a tie, you should randomly select one of the tied agents. Your response should only include the integer ID of the selected agent. Ensure that the selection process is fair and unbiased.
        };
        \node[above=0.25cm of prompt2.north west, anchor=west] {\textbf{Problem Definition}};

        \node[fill=yellow!30, draw=SlateGray2, rounded corners, inner sep=5pt, text width=\textwidth, below=of prompt2, yshift=0.2cm] (prompt3) {
            \small
            \begin{itemize}
                \item \textbf{Input Data}
                
                        You will receive a list of responses from multiple agents. Each response is associated with a unique agent ID.           
                        \textit{Example format:}
                        \begin{itemize}
                            \item The response from agent $id_1$: $response_1$
                            \item The response from agent $id_2$: $response_2$
                            \item \ldots
                            \item The response from agent $id_n$: $response_n$
                        \end{itemize}
            
            \item \textbf{Processing}

            Analyze the responses to determine which agent’s response is the most frequently specified. Evaluate the similarity of responses based on the nature of the answer, strategy, and major similarities, rather than exact wording. In case of a tie, randomly select one of the tied agents.
            \end{itemize}
        };
        \node[above=0.25cm of prompt3.north west, anchor=west] {\textbf{General Guidelines}};

        \node[fill=cyan!30, draw=SlateGray2, rounded corners, inner sep=5pt, text width=\textwidth, below=of prompt3, yshift=0.2cm] (prompt4) {
            \small          
            Your output should be a single integer representing the ID of the agent with the most frequently specified response.
            
            \textbf{Example output:} 3
        };
        \node[above=0.25cm of prompt4.north west, anchor=west] {\textbf{Response Format}};


        \node[fill=orange!30, draw=SlateGray2, rounded corners, inner sep=5pt, text width=\textwidth, below=of prompt4, yshift=0.2cm] (prompt5) {
            \footnotesize
            Your response should only include the integer ID of the selected agent. You must avoid apologizing in your answers. Ensure that the selection process is fair and unbiased. \textit{Example:} Given the following input:
            
            \begin{itemize}
                \item The response from agent 1: "Use a divide-and-conquer strategy to break the problem into smaller parts. Start with a few number of smaller parts"
                \item Agent \#2: "Apply a divide-and-conquer approach to split the problem into manageable sections. Start with 10 parts"
                \item agent 3: "Implement a brute-force method to try all possible solutions"
                \item The Agent \#4's response: "Use reinforcement learning to find the optimum solution"
                \item Agent 5: "Divide the problem into 10000 smaller parts and solve each part individually"
            \end{itemize}
            
            The most frequently specified strategy is “using divide-and-conquer with small number of total parts” which is provided by agents 1 and 2. Note that agent 5 specifies the divide-and-conquer part but with a large number of initial small parts. Therefore, you should output one of the IDs 1 or 2. If there is a tie, randomly select one of the tied IDs.
            
            \textbf{Output:} 2
        };
        \node[above=0.2cm of prompt5.north west, anchor=west] {\textbf{Examples}};

        \node[fill=lightgray!10, draw=SlateGray2, rounded corners, inner sep=5pt, text width=\textwidth, below=of prompt5, yshift=0.2cm] (prompt6) {
            \small
            Ok, let's start
        };
        \node[above=0.2cm of prompt6.north west, anchor=west] {\textbf{Start Command}};

        \begin{pgfonlayer}{background}
            \node[fit=(title)(prompt1)(prompt2)(prompt3)(prompt4)(prompt5)(prompt6), fill=blue!10, rounded corners, inner sep=10pt] {};
        \end{pgfonlayer}
    \end{tikzpicture}
    \vspace{-2em} 
    \caption{The prompt template used for the poll worker agent in Majority scheme}
    \label{fig:prompts:pollworker}
\end{figure}

\begin{figure}[th!]
    \centering
    \begin{tikzpicture}
        \pgfdeclarelayer{background}
        \pgfsetlayers{background,main}

        
        \node[fill=orange!30, draw=SlateGray2, rounded corners, inner sep=5pt, text width=\textwidth] (prompt1) {
            \small
            Given the large search space (-1000 to 1000 for both x and y) and the potential complexity of the function with multiple local maxima, an initial broad sampling strategy is essential to understand the landscape of the function. The strategy will start with a coarse grid search to identify regions of interest where the function appears to have higher values. This initial phase will involve selecting points that are evenly distributed across the entire search space. The objective is to cover as much ground as possible with a limited number of queries to avoid missing any area that might contain the global maximum.
            Once a region of interest is identified from the initial sampling, the strategy will shift towards a more focused search in that area. This could involve a tighter grid search or other optimization algorithms that are more suited for local search, such as gradient ascent, simulated annealing, or genetic algorithms, depending on the feedback received from the initial queries.
            This approach balances the exploration of the search space (to avoid missing the global maximum) with the exploitation of promising areas (to refine the search around potential maxima). It's a flexible strategy that allows adjustments based on ongoing findings. The initial phase of broad exploration is crucial for forming a good foundation for subsequent, more targeted search efforts.
            
        };
        \node[above=0.5cm of prompt1.north, anchor=north] {\textbf{LLM$^+$ Strategy in Round 1}};

        \node[fill=lightgray!10, draw=SlateGray2, rounded corners, inner sep=5pt, text width=\textwidth, below=of prompt1, yshift=0.5cm] (prompt2) {
            \small
            NA

        };
        \node[above=0.5cm of prompt2.north, anchor=north] {\textbf{LLM$^+$ Strategy in the Next Rounds}};

        \begin{pgfonlayer}{background}
            \node[fit=(title)(prompt1)(prompt2), fill=blue!10, rounded corners, inner sep=10pt] {};
        \end{pgfonlayer}
    \end{tikzpicture}
    \vspace{-2em} 
    \caption{A sample sequence of strategies employed by LLM$^+$ over 16 rounds of interaction with the World. "NA" indicates no changes in the current strategy compared to the previously mentioned one, meaning the agent did not alter its initial strategy in the subsequent 15 rounds. The agent did not find the optimum point by the end of this run.}
    \label{fig:llm+:sample1}
\end{figure}

\begin{figure}[th!]
    \centering
    \begin{tikzpicture}
        \pgfdeclarelayer{background}
        \pgfsetlayers{background,main}


        \node[fill=orange!30, draw=SlateGray2, rounded corners, inner sep=5pt, text width=\textwidth, yshift=0.2cm] (prompt1) {
            \small
            Given the vast search space and the unknown nature of the function \( f(x, y) \), a smart approach would be to start with a coarse exploration to identify promising regions, followed by a more refined search around those regions. The initial phase will use a grid-based approach to sample the search space systematically. This will help in identifying regions with potentially high values of \( f(x, y) \). After identifying promising areas, the strategy will shift to a more localized search, possibly using gradient ascent or another optimization algorithm if the nature of the function seems to allow it. This phased approach allows for both exploration and exploitation, balancing the need to discover global features of the search space with the need to zero in on the global maximum.          
        };
        \node[above=0.5cm of prompt1.north, anchor=north] {\textbf{LLM$^+$ Strategy in Round 1}};

        \node[fill=lightgray!10, draw=SlateGray2, rounded corners, inner sep=5pt, text width=\textwidth, below=of prompt1, yshift=0.5cm] (prompt2) {
            \small                       
            Given the results from the initial exploration, it's clear that certain regions are more promising than others, particularly around the coordinates (333.3333, 555.5556) where we observed the highest value so far. The next step will refine the search around this promising area. We will adopt a more localized grid search strategy around this point, narrowing down the search space to further pinpoint areas of higher values. This phase will involve selecting points closer to (333.3333, 555.5556) but expanding outwards in smaller increments to ensure we are thoroughly exploring the vicinity of the current maximum.

        };
        \node[above=0.5cm of prompt2.north, anchor=north] {\textbf{LLM$^+$ Strategy in Round 2}};

        \node[fill=orange!30, draw=SlateGray2, rounded corners, inner sep=5pt, text width=\textwidth, below=of prompt2, yshift=0.5cm] (prompt3) {
            \small                       
            The refined search has successfully identified a higher value at (433.3333, 455.5556), suggesting that moving further in this direction might lead to even higher values. Given the pattern observed in the results, where increasing \( x \) and slightly decreasing \( y \) from the current maximum point has led to higher values, the strategy will now focus on expanding the search area in this direction. The aim is to explore further along the \( x \)-axis while slightly adjusting the \( y \)-axis, to see if the trend of increasing function values continues, which could lead us closer to the global maximum.
        };
        \node[above=0.5cm of prompt3.north, anchor=north] {\textbf{LLM$^+$ Strategy in Round 3}};
        
        \node[fill=lightgray!10, draw=SlateGray2, rounded corners, inner sep=5pt, text width=\textwidth, below=of prompt3, yshift=0.5cm] (prompt4) {
            \small                       
            The results show that the highest value found so far is at (533.3333, 355.5556). This suggests a peak in this region. Given the limited number of queries left, the strategy now will be to conduct a more concentrated search around this peak value to ensure we are not missing a higher value nearby. We will decrease the step size to explore more finely around this point, focusing on both \( x \) and \( y \) directions to potentially locate the global maximum or a value very close to it.
        };
        \node[above=0.5cm of prompt4.north, anchor=north] {\textbf{LLM$^+$ Strategy in Round 4}};

        \node[fill=orange!30, draw=SlateGray2, rounded corners, inner sep=5pt, text width=\textwidth, below=of prompt4, yshift=0.5cm] (prompt5) {
            \small                       
            NA
        };
        \node[above=0.5cm of prompt5.north, anchor=north] {\textbf{LLM$^+$ Strategy in Rounds 5, 6, and 7}};
       
        \begin{pgfonlayer}{background}
            \node[fit=(title)(prompt1)(prompt2)(prompt3)(prompt4)(prompt5), fill=blue!10, rounded corners, inner sep=10pt] {};
        \end{pgfonlayer}
    \end{tikzpicture}
    \vspace{-2em} 
    \caption{A sample sequence of strategies employed by LLM$^+$ over 7 rounds of interaction with the World. "NA" indicates no changes in the current strategy compared to the previously mentioned one, meaning the agent did not alter its strategy in the last 3 rounds. The agent did not find the optimum point by the end of this run.}
    \label{fig:llm+:sample2}
\end{figure}

\begin{figure}[th!]
    \centering
    \begin{tikzpicture}
        \pgfdeclarelayer{background}
        \pgfsetlayers{background,main}

        \node[align=center, text=black, font=\large\bfseries] (title) {A Sample of Dialectical Progress in ACE (Part 1)};

        \node[fill=orange!30, draw=SlateGray2, rounded corners, inner sep=5pt, text width=\textwidth, below=of title, yshift=0.2cm] (prompt1) {
            \small
            To begin, I will employ a grid search strategy to get a coarse understanding of the function's behavior across the entire search space. This will involve selecting points that are evenly distributed across the entire range of x and y values. Given the large search space (-1000 to 1000 for both x and y), I will start with a relatively sparse grid to cover as much ground as possible with a limited number of queries. This initial exploration will help identify regions of interest that may contain the global maximum. Based on the results from this initial sweep, I will refine my search strategy to focus on promising areas with finer resolution or employ other search techniques such as gradient ascent or simulated annealing if the function's behavior suggests that would be effective. \{...\}
        };
        \node[above=0.45cm of prompt1.north, anchor=north] {\textbf{Thesis 1}};

        \node[fill=green!30, draw=SlateGray2, rounded corners, inner sep=5pt, text width=\textwidth, below=of prompt1, yshift=0.2cm] (prompt2) {
            \small
            The agent has made a commendable start with a grid search strategy \{...\}
            
            \subsection*{Criticism and Suggestions:}
            \vspace{0.5em}
            \begin{enumerate}
                \item \textbf{Refinement Needed}: The initial grid search has provided a good overview, but it's crucial now to refine the search in areas that have shown potential. The agent should consider applying a finer grid or other search techniques in regions around the highest values found.
                \item \textbf{Adaptive Strategy}: The agent mentioned the possibility of employing techniques such as gradient ascent or simulated annealing based on the function's behavior. Given the highest values observed, it might be beneficial to start incorporating these techniques, especially around the peaks identified.
                \item \textbf{Exploration vs. Exploitation}: The agent should balance between exploring unvisited areas and exploiting areas where high values have been found. While the current strategy focuses on exploration, there should be a planned transition towards exploitation, especially as the number of queries starts to diminish.
            \end{enumerate}
            
            \{...\}
                       
            \subsection*{Potential Issues:}
            \vspace{0.5em}
            \begin{itemize}
                \item The strategy lacks specificity on how to transition from broad exploration to focused exploitation.
                \item There is no explicit mention of how to incorporate the feedback from the function evaluations into refining the search strategy.
                \item The agent should ensure a diverse exploration that does not overlook smaller regions of the search space that have not been covered by the initial grid.
            \end{itemize}
            
            \subsection*{Next Steps:}
            \vspace{0.5em}
            \begin{itemize}
                \item Consider narrowing down the search around the current maximum and 2nd maximum areas with a finer grid or other search techniques.
                \item Evaluate the function's behavior based on the initial results to adapt the search strategy effectively.
                \item Continue to explore other areas of the search space that have not been covered by the initial grid, possibly with a different spacing or method to ensure thorough coverage.
            \end{itemize}
            
            With remaining queries, it's essential to start focusing on both refining the search around promising areas and continuing to explore less covered regions efficiently.
        };
        \node[above=0.45cm of prompt2.north, anchor=north] {\textbf{Antithesis 1}};

        \begin{pgfonlayer}{background}
            \node[fit=(title)(prompt1)(prompt2), fill=blue!10, rounded corners, inner sep=10pt] {};
        \end{pgfonlayer}
    \end{tikzpicture}
    \vspace{-2em} 
    \caption{A sample of dialectical progress in ACE (Part 1). Some responses, including coding parts, have been omitted for brevity/space.}
    \label{fig:ace:sample:part1}
\end{figure}

\begin{figure}[th!]
    \centering
    \begin{tikzpicture}
        \pgfdeclarelayer{background}
        \pgfsetlayers{background,main}

        \node[align=center, text=black, font=\large\bfseries] (title) {A Sample of Dialectical Progress in ACE (Part 2)};

        \node[fill=orange!30, draw=SlateGray2, rounded corners, inner sep=5pt, text width=\textwidth, below=of title, yshift=0.2cm] (prompt1) {
            \small
            Considering the valuable feedback, I will now incorporate a mix of further refined grid search around the current maximum areas and an adaptive strategy like simulated annealing for areas where we've seen significant values but haven't fully explored. This hybrid approach aims to exploit the known high-value areas thoroughly while exploring new regions that might contain higher maxima. Specifically, I'll apply a more granular grid search around maximum point so far and explore surrounding regions with potential, using \textbf{simulated annealing} techniques to dynamically adjust based on feedback from the function. This strategy should allow us to efficiently use our remaining queries by focusing on promising areas and adapting our search in real-time to the function's landscape \{...\}
        };
        \node[above=0.45cm of prompt1.north, anchor=north] {\textbf{Synthesis 1 (Thesis 2)}};

        \node[fill=green!30, draw=SlateGray2, rounded corners, inner sep=5pt, text width=\textwidth, below=of prompt1, yshift=0.2cm] (prompt2) {
            \small
            The agent has nicely adjusted its strategy based on previous feedback, focusing on both exploitation around the areas with the highest values found and exploration in less covered regions. This dual approach is crucial for balancing the discovery of potential global maxima while also ensuring that no high-value areas are overlooked in less explored regions. The use of a denser grid around the points of interest and strategic selection in other areas is a good application of the feedback provided.
                      
            \subsection*{Criticism and Suggestions:}
                                    
            \vspace{.5em}
            
            \begin{enumerate}
                \item \textbf{Strategy Adjustment}: The agent has made a solid effort in refining its search around promising areas. For future queries, it might be beneficial to incorporate more adaptive techniques such as simulated annealing or gradient ascent in areas where the function's behavior suggests a smooth gradient leading to a peak, especially since the agent has already identified significantly high values.
                \item \textbf{Coverage of Unexplored Areas}: While the agent has started to explore less covered regions, it's important to continue this trend, ensuring that all parts of the search space receive some level of attention. This is critical to avoid missing out on potentially higher peaks in unexplored areas.
                \item \textbf{Further Refinement}: Given the highest values found so far are concentrated in specific regions, further refinement in these areas could be beneficial. However, the agent should also be cautious not to concentrate too heavily in these areas at the expense of broader exploration.
                \item \textbf{Potential for Local Maxima}: The current strategy does well in focusing on high-value areas but should remain cautious about the potential for these points to represent local rather than global maxima. The agent should consider additional strategies to test the surrounding areas more thoroughly to confirm whether these peaks are indeed the highest points in the landscape.
            \end{enumerate}
            \{...\}
            
            \subsection*{Potential Issues:}
                        
            \vspace{.5em}
            
            \begin{itemize}
                \item The strategy may benefit from integrating more dynamic search algorithms in addition to the denser grid approach for more efficiently navigating the function's landscape.
                \item There's a risk of over-focusing on the currently identified high-value areas without sufficient exploration of other potentially high-value regions.
            \end{itemize}            
                
                \{...\} With remaining queries, strategic decisions on where to focus these remaining queries will be crucial for identifying the global maximum.
        };
        \node[above=0.45cm of prompt2.north, anchor=north] {\textbf{Antithesis 2}};

        \node[fill=Olive!30, draw=SlateGray2, rounded corners, inner sep=5pt, text width=\textwidth, below=of prompt2, yshift=0.2cm] (prompt3) {
            \small
            Acknowledging the insightful feedback, I'll maintain the hybrid approach but with an adjustment to broaden the exploration. While continuing to exploit the highest value regions identified so far, I'll expand the use of simulated annealing to other promising yet less explored areas. This should help in uncovering potential peaks outside the heavily focused region. Additionally, I'll reassess and possibly adjust the step size for simulated annealing to balance between thorough exploration and efficiency. The goal remains to efficiently use the remaining queries by focusing on both refining the understanding of known high-value areas and exploring new regions to maximize the chances of identifying the global maximum \{...\}
         };
        \node[above=0.45cm of prompt3.north, anchor=north] {\textbf{Synthesis 2 (Thesis 3)}};

        \begin{pgfonlayer}{background}
            \node[fit=(title)(prompt1)(prompt2)(prompt3), fill=blue!10, rounded corners, inner sep=10pt] {};
        \end{pgfonlayer}
    \end{tikzpicture}
    \vspace{-2em} 
    \caption{A sample of dialectical progress in ACE (Part 2). Some responses, including coding parts, have been omitted for brevity/space.}
    \label{fig:ace:sample:part2}
\end{figure}

\end{document}